\documentclass[10pt,twocolumn,letterpaper]{article}
\makeatletter
\providecommand{\@LN}[2]{}
\makeatother

\usepackage[pagenumbers]{cvpr} 

\usepackage{algorithm}
\usepackage{algpseudocode}
\usepackage{algorithmicx}
\usepackage[dvipsnames]{xcolor}
\usepackage{multirow}
\usepackage{tabularray}
\usepackage{graphicx}
\usepackage{booktabs}
\usepackage{amsmath}
\usepackage{subcaption}

\usepackage{amsthm}
\newtheorem*{remark}{Remark}
\usepackage{afterpage}
\usepackage{dblfloatfix}
\usepackage{placeins}

\usepackage{multirow}
\usepackage{makecell}

\definecolor{cvprblue}{rgb}{0.21,0.49,0.74}
\usepackage[pagebackref,breaklinks,colorlinks,citecolor=cvprblue]{hyperref}

\def\paperID{12570} 
\def\confName{CVPR\xspace}
\def\confYear{2025\xspace}

\title{Simulator HC: Regression-based Online Simulation of Starting Problem-Solution Pairs for Homotopy Continuation in Geometric Vision}

\author{Xinyue Zhang$^1$ \and Zijia Dai$^1$ \and Wanting Xu$^1$ \and Laurent Kneip$^1$ \\
\and 
\hspace*{12pt}
$^{1}$Mobile Perception Lab, ShanghaiTech University, China}

\begin{document}
\maketitle
\begin{abstract}
  While automatically generated polynomial elimination templates have sparked great progress in the field of 3D computer vision, there remain many problems for which the degree of the constraints or the number of unknowns leads to intractability. In recent years, homotopy continuation has been introduced as a plausible alternative. However, the method currently depends on expensive parallel tracking of all possible solutions in the complex domain, or a classification network for starting problem-solution pairs trained over a limited set of real-world examples. 
  Our innovation lies in a novel approach to finding solution-problem pairs, where we only need to predict a rough initial solution, with the corresponding problem generated by an online simulator.
  Subsequently, homotopy continuation is applied to track that single solution back to the original problem. We apply this elegant combination to generalized camera resectioning, and also introduce a new solution to the challenging generalized relative pose and scale problem. As demonstrated, the proposed method successfully compensates the raw error committed by the regressor alone, and leads to state-of-the-art efficiency and success rates.
\end{abstract}
\vspace{-0.5cm}    
\section{Introduction}
\label{sec:intro}
The solution of geometric camera calibration problems is a crucial step in many Structure-from-Motion (SfM)~\cite{hartley2004multiple,snavely06,snavely08,schonberger16}, Visual Odometry (VO)~\cite{nister04}, and visual Simultaneous Localization And Mapping (SLAM)~\cite{murartal15} frameworks. Solvers developed over the years address camera resectioning~\cite{kneip11,hesch11,zheng13,kneip2014upnp}, two-view relative pose~\cite{nister2004efficient,stewenius06,kneip12}, pose estimation from lines~\cite{elqursh11,miraldo18}, partially calibrated cameras~\cite{zheng16}, generalized cameras~\cite{stewenius05,kneip2014upnp,ventura15,miraldo18}, rolling shutter cameras~\cite{saurer15,dai16}, and many other scenarios. The problems often appear in the form of a polynomial equation system, and special techniques from the field of algebraic geometry have been used to solve them. Once an efficient solver is found, it is typically embedded into a random sampling and consensus scheme~\cite{fischler81,raguram13} in order to gain robustness in the presence of outliers. Hence, there is a requirement for such solvers to be efficient and embeddable into iterative schemes.

The dominant approaches to efficiently solve systems of polynomial equations are given by the Gr\"obner basis method and polynomial resultants~\cite{cox98,sturmfels02}. Based on the Gr\"obner basis theory, Stew{\'e}nius et al.~\cite{stewenius06} and Kukelova et al.~\cite{kukelova08} demonstrate that the calculations required to obtain a minimal set of ideal generators can be effectuated efficiently at the hand of a fixed elimination template. Kukelova et al.~\cite{kukelova08} in particular propose a solver generator to automatically discover such elimination templates. Later on, sparse resultant-based solvers~\cite{emiris2012general} have been demonstrated as a powerful alternative sometimes leading to smaller elimination templates.

While polynomial elimination techniques have led to significant progress in the solution of geometric vision problems, the methods are limited as the complexity of the template search grows uncontrollably with the degree and dimensionality of the problem at hand. Furthermore, owing to their failure to respect inequality constraints and rule out practically infeasible solutions, high degrees or dimensionalities naturally lead to an elevated number of algebraically possible solutions and---assuming that it can be found at all---large elimination templates~\cite{nister06,zheng13,zheng16,duff19,duff20,zhao20}. This in turn causes problems in terms of numerical stability~\cite{zhao20}. In recent years, a different solution strategy has therefore gained popularity: \textit{Homotopy Continuation} (HC)~\cite{verschelde10,bates13,duff18}. The method proceeds by starting from a known problem-solution pair, and then tracks its roots to the target problem at hand during a step-wise interpolation of polynomial coefficients. The method has been used to successfully solve challenging tri-focal relative pose problems~\cite{fabbri20}. While promising, HC remains expensive as it generally requires the parallel continuation of many roots in the complex domain in order to identify all real solutions to a target problem, followed up by disambiguation.

Recently, Hruby et al.~\cite{hruby2022learning} propose an efficient, learning-based extension to the pure HC paradigm, which serves as our primary motivation. They propose the introduction of a classifier to pick a good problem-solution pair trained over a large set of known pairs obtained from SfM. This enables a highly efficient application of HC, as only a single root of interest needs to be continued. While still presenting a margin for improvement in terms of success rate, the method is demonstrated to have unprecedented computational efficiency on the challenging three-view-four-point problem.

The method proposed in this work is similar to the work of Hruby et al.~\cite{hruby2022learning} in that it leverages a learning-based approach to produce a starting problem-solution pair that enables single-root-tracking. However, we note a couple of important differences:
%
%
\begin{itemize}
  \item We do not train a classifier on a fixed set of problem-solution pairs taken from a real-world dataset. Instead, we propose to employ a solution regression network trained over an arbitrarily large set of simulated input correspondences for the considered geometric problem. The network is hence not limited by a finite set of candidate problems, and aims at being a general solution approximator for any instance of the polynomial problem. What's more, learning a solution is much easier than learning a proper problem-solution pair.
  \item The bridge to HC is formed by appending an online simulator that uses the input correspondences and the predicted solution to generate a complete, consistent problem-solution pair. Given only moderate regression accuracy, the produced problem-solution pair is sufficiently close to the original problem for successful single-root tracking via HC.
  \item We argue that the prediction of a single solution (or the unambiguous classification of a single problem-solution pair) from a minimal set may not always be possible as for complicated geometric problems, there may indeed be multiple geometrically feasible solutions. We therefore add one additional correspondence, which leads to a performance gain during both the regression and the continuation stage.
\end{itemize}

\begin{figure*}[h]
    \centering
    \includegraphics[width=0.9\linewidth]{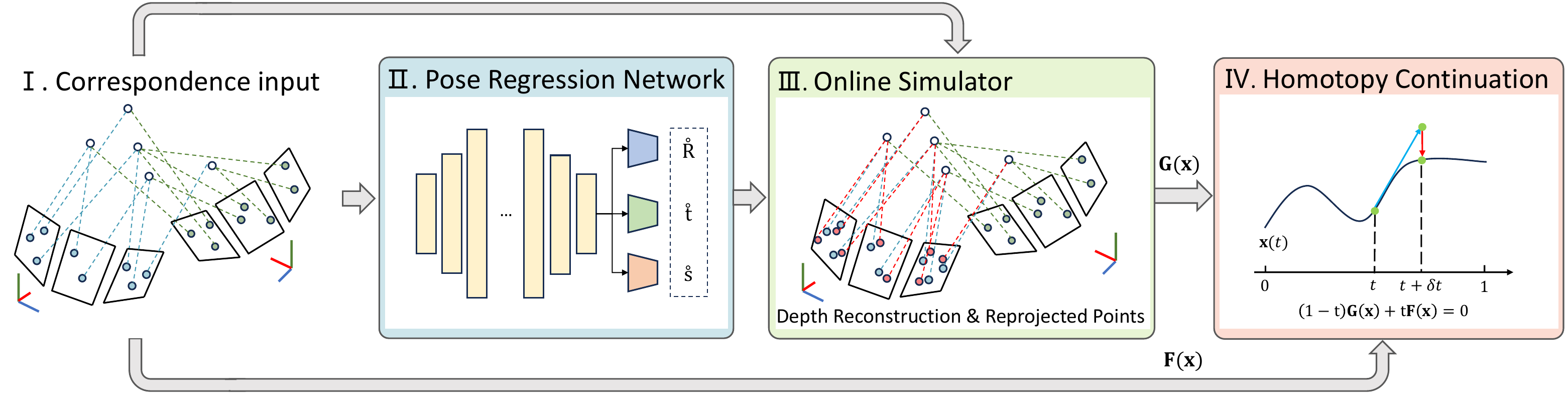}
    \caption{Overview of the proposed geometric problem solution scheme. Given input correspondences, a regression network is utilized to approximate a solution. A subsequent online simulator generates a new set of correspondences that is consistent with the regression output. The obtained problem-solution pair is finally used to bootstrap homotopy continuation. The final solution is found efficiently by tracking a single root.}
    \label{fig:overview}
    \vspace{-0.5cm}
\end{figure*}

The newly proposed \textit{simulate-and-solve} paradigm is summarized in Figure~\ref{fig:overview}. We demonstrate that this strategy can lead to high success-rate and efficient CPU-based solvers for two problems: generalized camera resectioning, and generalized relative pose and scale. Owing to its challenging nature, the latter problem has thus far only seen an optimization-based solution~\cite{kneip2016generalized}. The solvers are compared against a purely learning-based alternative, a regressor followed by simple local refinement, as well as the exhaustive application of homotopy continuation over all roots. The proposed method is efficient, generalizable, and highly successful, and should thus be considered as an interesting alternative for the ongoing development of complicated geometric problem solvers.

\section{Literature Review}\label{sec:Literature}

\noindent\textbf{Polynomial elimination theory:} Our contribution can be regarded as a novel alternative to traditional polynomial elimination theory~\cite{cox98,sturmfels02}. Stewenius et al.~\cite{stewenius06} manually derive one of the first Gr\"obner basis solvers in the field of computer vision. Automatic solver generators and important extensions improving performance and numerical accuracy are later introduced by Kukelova et al.~\cite{kukelova08} and Larsson et al.~\cite{larsson17a,larsson17b,larsson17c,larsson18}. Zheng and Kneip propose the use of the parametric Sylvester resultant~\cite{zheng16}, while Hesch and Roumeliotis make use of the Macaulay resultant~\cite{hesch11}. The more general use of sparse resultants as a powerful alternative to the Gr\"obner basis method is analyzed by Emiris~\cite{emiris2012general}, Heikkil{\"a}~\cite{heikkila17}, and---most recently---Bhayani et al.~\cite{bhayani20}. While having lead to significant progress, complicated geometric problems with as many as 64~\cite{stewenius05}, 272~\cite{nister06}, 81~\cite{zheng13} (before symmetry-removal), 210~\cite{zheng16}, $>$10000~\cite{duff19,duff20}, and 512~\cite{zhao20} solutions remain hard or impossible to be solved by polynomial elimination. The latter work, in particular, presents one of the largest successful elimination templates in the literature (96413$\times$96879) by making use of octuple machine precision.
Recently Evgenity et al.~\cite{martyushev2022optimizing} developed a new method to reduce the size of the elimination template.
Yet, for a polynomial system, a large number of solutions in practice still leads to reduced computational efficiency.

\noindent\textbf{Homotopy continuation:} Homotopy continuation implementations exist as independent programs~\cite{verschelde10} or as part of Macaulay2~\cite{leykin11}, Bertini~\cite{bates13}, and Julia~\cite{breiding18}. Duff et al.~\cite{duff18} study the application of the method to geometric vision problems, and several related studies focus specifically on the tri-focal relative pose problem~\cite{haralick94,holt95,quan06,aholt14,kileel17}. However, off-the-shelf implementations are often too general and not optimized for real-time applications. Progress is recently achieved by Fabbri et al.~\cite{fabbri20}, who propose an efficient custom C++ implementation of the Macaulay2 solver~\cite{leykin11} for a tri-focal relative pose problem. More recently, Ding et al.~\cite{ding2023minimal} propose a fast GPU-based implementation of homotopy continuation~\cite{chien2022gpu,chien2022parallel} for a minimal solution to the three-view-four-point problem which has the same performance as Julia~\cite{breiding18} but is significantly faster. The method leverages massive parallelization to efficiently track all solutions of a small, fixed set of starting problem-solution pairs. Our contribution can be regarded as a problem-solution pair initialization scheme similar to Hruby et al.~\cite{hruby2022learning} for successful single solution tracking.

\noindent\textbf{Learning for polynomial problem solving:} One of the first methods to successfully employ learning in polynomial solvers is presented by Xu et al.~\cite{xu19}, who proposed to increase the numerical accuracy of polynomial elimination templates by learning the best permutation in permutation-invariant polynomial problems. Recently, Hruby et al.~\cite{hruby2022learning} propose the addition of a classifier for initial problem-solution pair picking in homotopy continuation. In recent years, several works have proposed to directly apply neural networks to regress the pose of a camera, either directly from images~\cite{kendall2017geometric,sarlin2021back} or---aiming at replacing traditional algebraic-geometric solvers---from correspondence coordinates~\cite{sheffer2020pnpnet}. While such networks use algorithm unrolling~\cite{monga2021algorithm} and differential optimization~\cite{amos2017optnet} for accurate end-to-end trainability, a successful application to more complicated problems remains yet to be demonstrated. Regressors can be regarded as general non-linear function approximators, and the challenge of using them to produce highly accurate solutions is easily explained by the potential discontinuity of root locations for continuous variations of input polynomial coefficients. Our contribution can be regarded as a new post-regression refinement scheme for polynomial problems enabling successful application to complicated problems for which it is hard to apply traditional polynomial solvers.
\section{Polynomial Solver: Homotopy Continuation}
\begin{figure}
    \centering
    \begin{subfigure}{0.45\linewidth} 
    \includegraphics[width=\linewidth]{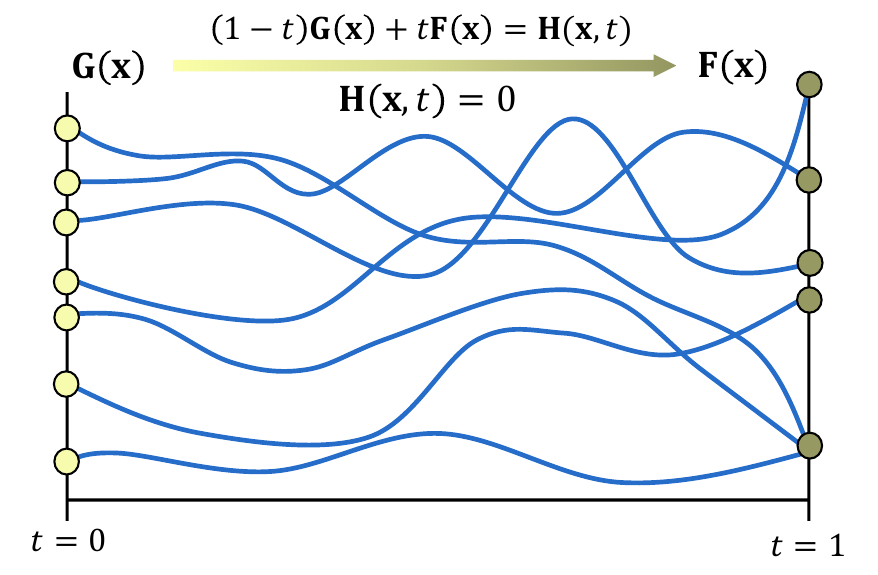}
    \caption{HC multiple solution curves}
    \label{sub-fig:HCntrack}
    \end{subfigure}
    \qquad
    \begin{subfigure}{0.45\linewidth} 
    \includegraphics[width=\linewidth]{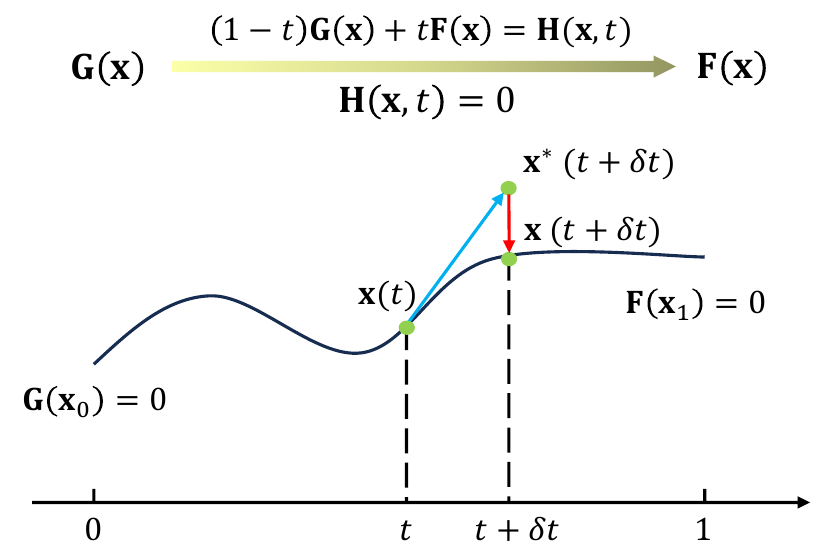}
    \caption{HC solution tacking}
    \label{sub-fig:HCsingletrack}
    \end{subfigure}
    \caption{Graphical illustration on homotopy continuation(HC) solution curves tracking. 
    For a polynomial system with multiple solutions, each solution curve can be tracked independently as shown in \figureautorefname~\ref{sub-fig:HCntrack}. For each solution curve tracking, HC utilizes a ``prediction-correction'' scheme to approximate one of the solutions for the target system. 
    \vspace{-0.5cm}}
    \label{fig:HC_visual}
\end{figure}
We start with a brief summary of HC.
We refer to an instance of a certain polynomial system with specific coefficients as a \textit{problem}.
HC is a powerful tool that can numerically find the solutions of a problem. 
Each solution is found by tracking a curve independently, as illustrated in \figureautorefname~\ref{sub-fig:HCntrack}. 
When tracking the solution curves, the method solves a series of problems of similar form. 
It starts from a trivial problem whose solutions are simply known, and then gradually deforms that problem to the target problem that we are interested in but for which we do not know the solutions. It is hoped that---by continuing the known solutions throughout the problem sequence---they can lead to the solutions of the target, input problem. Two questions arise: 1) How to track the solution curves?, and 2) What is a good starting problem leading to the solutions we desire?

The answer to the first question is given by HC. Formally, suppose we have a polynomial system $\mathbf{F}(\mathbf{x}) = \mathbf{0}$ and we are interested in its solutions.
Suppose further that we have a polynomial system of similar form $\mathbf{G}(\mathbf{x})=\mathbf{0}$ whose solutions are known.
HC solves a series of polynomial systems of the form 
\begin{equation}
    \mathbf{H}(\mathbf{x}, t) = (1-t) \mathbf{G}(\mathbf{x}) + t \mathbf{F}(\mathbf{x}) , \quad t \in [0, 1].
    \label{eq:H(x,t)}
\end{equation}
As illustrated in \figureautorefname~\ref{sub-fig:HCsingletrack}, each $\mathbf{H}(\mathbf{x}, t) = \mathbf{0}$ is solved while sliding $t$ from 0 to 1. For $t=0$, the solution is known as $\mathbf{H}(\mathbf{x}, 0) = \mathbf{G}(\mathbf{x})$. A solution of $\mathbf{F}(\mathbf{x})=\mathbf{0}$ is finally obtained as for $t=1$, $\mathbf{H}(\mathbf{x}, 1) = \mathbf{F}(\mathbf{x})$.
The problems are not actually solved but the solutions are simply tracked from problem to problem. Taking the partial derivative to both sides of $\mathbf{H}(\mathbf{x}, t)$, the solution curve $\mathbf{x}(t)$ can be formulated as the following ordinary differential equation
\begin{equation}
    \mathbf{J}_{\mathbf{H}}(\mathbf{x}) \frac{\partial \mathbf{x}}{\partial t} + \frac{\partial \mathbf{H}(\mathbf{x}, t)}{\partial t} = \mathbf{0}
    \iff
    \frac{\partial \mathbf{x}}{\partial t} = -{\mathbf{J}_{\mathbf{H}}(\mathbf{x})}^{\dagger} \frac{\partial \mathbf{H}(\mathbf{x}, t)}{\partial t}. \label{eq:ODE}
\end{equation}
A ``prediction-correction'' scheme---as widely utilized in numerical analysis---is typically employed to trace the solution curve $\mathbf{x}(t)$. 

\noindent \textbf{Prediction.}
Denote the stepsize of $t$ as $\delta t$. Euler's predictor can be used to update along the tangent direction
\begin{equation}
    \mathbf{x}_{i+1} = \mathbf{x}_{i} + \delta t \frac{\partial \mathbf{x}}{\partial t}\bigg |_{\mathbf{x}_i, t_i}.
\end{equation}
Practically speaking, a higher-order generalized method is commonly used for better accuracy and stability with larger stepsize.

\noindent \textbf{Correction.} Gauss-Newton steps are adopted to correct the prediction.
In each iteration of Gauss-Newton's method, the solution is updated by
\begin{equation}
    \mathbf{x}_j = \mathbf{x}_{j-1} - {\mathbf{J}_{\mathbf{H}}(\mathbf{x}_{j-1})}^{\dagger} \mathbf{H}(\mathbf{x}_{j-1},t),
\end{equation}
where $\mathbf{J}_{\mathbf{H}}^\dagger$ is the pseudo inverse of the jacobian $\mathbf{J}_{\mathbf{H}}$, and $t = t_i + \delta t$.
\figureautorefname~\ref{sub-fig:HCsingletrack} illustrates a 1D solution curve tracking, where the blue arrow represents the prediction update and the red arrow represents the correction update.

The remainder of this paper---our key contribution---addresses the second question of how to find a good starting system. Traditional methods (e.g. Ab initial~\cite{fabbri20}, monodromy \cite{duff18, breiding18}) rely on algebraic geometry theories developed for coefficient-parameter homotopies, and tracking all roots from a small set of fixed, known problem-solution pairs in the complex domain may often enable a solution~\cite{sommese2005numerical}. However, tracking all the isolated paths is expensive and in practice multiple solutions cost more time to select the best. A recent study proposes the use of a classifier to pick a good starting pair with as high as possible success probability upon tracking only a single root~\cite{hruby2022learning}, which is much faster.
Similarly, in this work, we propose a novel geometry-informed strategy that combines a neural regressor with an online simulator to produce a problem-solution pair and then track a single root by HC. Even on challenging problems for which no traditional solver has been found, it can find the solution with a success rate of over $95\%$.

\begin{remark}[Homotopy Types]
    Since being introduced in 1980s, a variety of homotopies have been developed to serve the same purpose: how to solve the problem exactly as expected. The representative one would be coefficient-parameter homotopy. Its formulation provides the insight that if we can find all the solutions to a general coefficient-parameter polynomial, then we can find all the other members of that polynomial family.
    In fact, as pointed out in~\cite{sommese2005numerical}, in engineering problems (e.g. geometry vision problems) we are dealing with natural coefficient-parameter polynomials, that is coefficients of polynomials derived from visual features.
    Since the scope of this paper lies on a novel starting problem-solution pair finding and it is not highly dependent on the path tracking, 
    we kindly refer the reader to \cite{sommese2005numerical} for an in-depth discussion of other homotopy types.
    As shown in the later applications, straight segment homotopies provide near-perfect success rates when working with our simulator. We also want to note that all the other homotopy types could be equally applied in conjunction with our simulator.
\end{remark}

\section{Find a Good Starting Solution: Regression Network}
\label{sec:network}
Following recent developments in deep learning, multiple attempts have been made to, for example, learn camera pose using a neural network \cite{kendall2017geometric,sarlin2021back,sheffer2020pnpnet}.
The first two methods extract features from images and points and use them in a regression network to predict the pose. The third method is a more preliminary approach and aims at predicting pose directly from geometric correspondences, which is also the aim of the present work.
The bottleneck in regressing pose from correspondences arises from two concerns: 1) For simple geometric problems, we often have traditional methods that are efficient and accurate, and 2) For hard geometric problems, regression networks demonstrate a poor ability to predict accurate solutions.
Our key insight is that while the predicted pose may not be sufficiently accurate, it can still serve as a good starting solution for HC.

Our pose regression networks are designed with multiple 1D convolutional layers and each layer is followed by batch normalization and ReLU activations.
The loss function is given by a combination of relative pose errors with learnable weights~\cite{kendall2017geometric}. 

\section{Find a Consistent Starting Problem: Online Simulator}
\label{sec:Offline-Simulator}
We recall the main idea of HC. In order to continue a solution to the target problem, an appropriate starting problem with a known solution is required.
Given the approximate solution predicted by the pose regression network,
our goal is to derive a consistent polynomial system for which one of the solutions matches the predicted one. We denote this step an \textit{online simulator}, and---though problem specific---effective realizations can almost always be found by simple geometric rules. For example, in camera resectioning, one can simply use the input 3D points and reproject them into a camera view under the predicted pose. This will lead to new 2D points, and thus a new set of correspondences that is perfectly consistent with the predicted solution, i.e. our sought starting problem-solution pair. While not yet made use of in the present work, it is worthwhile noting that---owing to their algebraic simplicity---such an online simulator often appears to be a differentiable module.

\begin{figure}
    \centering
    \subcaptionbox{{\footnotesize Generalized Absolute Pose} \label{sub-fig:upnp}}[0.45\linewidth]{ 
        \includegraphics[width=\linewidth]{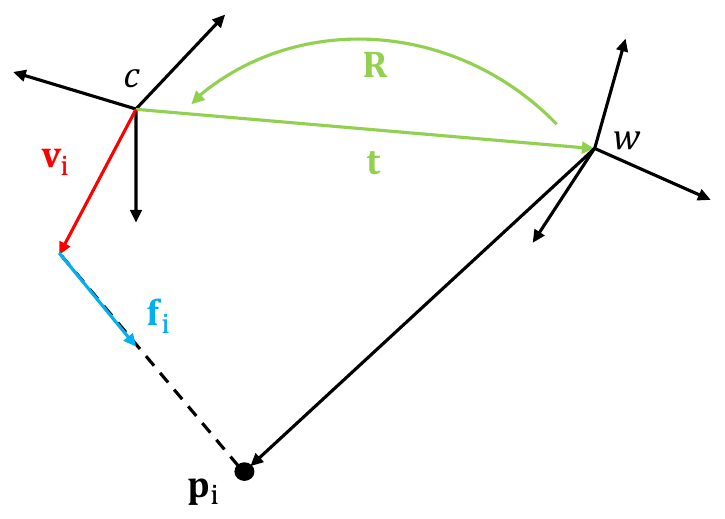}
    }\hfill 
    \subcaptionbox{{\footnotesize Generalized Relative Pose and Scale} \label{sub-fig:grps}}[0.45\linewidth]{ 
        \includegraphics[width=\linewidth]{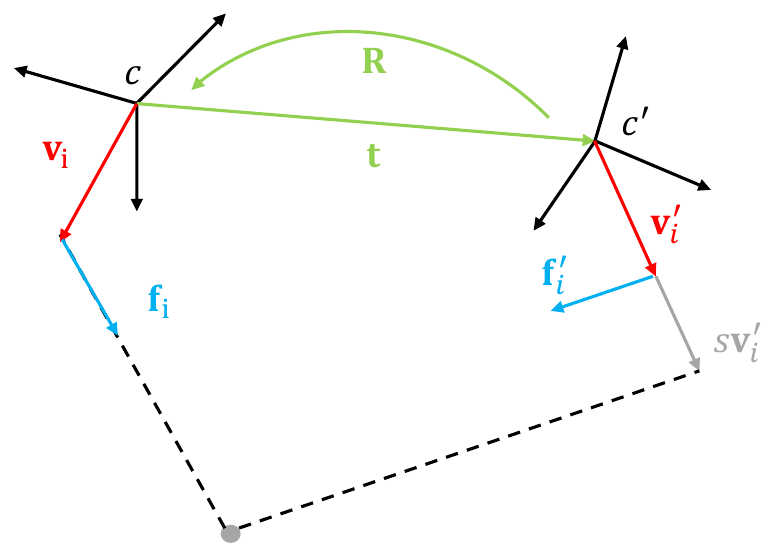} 
    }
    \caption{The geometry of the generalized camera problems we propose to solve by online simulator HC. $w$ represents the world frame, and $c$ and $c'$ represent camera frames in different views respectively.}
    \label{fig:camera_models}
    \vspace{-0.6cm}
\end{figure}

\section{Generalized Perspective-n-Points}
Our first concrete, practical application targets generalized camera pose estimation from 2D-3D correspondences. We use the Unified Perspective-n-Point (UPnP) algorithm~\cite{kneip2014upnp} which formulates a polynomial system corresponding to the first-order optimality conditions of the sum of squared object space distances. The method can be used to solve both central and non-central camera resectioning problems from an arbitrary number of points (including minimal). It can be efficiently solved by the Gr\"{o}bner Bases method since---after variable elimination---it only involves $4$ unknowns. In this section, we solve UPnP by our newly proposed method and compare it against conventional approaches, a raw regression network, as well as a parallel HC implementation tracking all roots.

\subsection{Problem Formulation}
Consider a 3D-2D generalized correspondence $(\mathbf{p}_i, \mathbf{f}_i, \mathbf{v}_i)$ originating from the non-central camera resectioning model illustrated in \figureautorefname~\ref{sub-fig:upnp}, 
where $\mathbf{p}_i\in\mathbb{R}^{3}, \mathbf{f}_{i}\in\mathbb{S}^{2}$ and $\mathbf{v}_{i}\in\mathbb{R}^{3}$ denote the 3D world point, image ray and ray origin, respectively.
The correspondence follows an underlying generalized absolute pose transformation depending on the rotation $\mathbf{R}\in SO(3)$, the position $\mathbf{t}\in\mathbb{R}^{3}$ and an unknown latent depth $\alpha_i$ as follows:
\begin{equation}
\alpha_i\mathbf{f}_i+\mathbf{v}_i=\mathbf{R}\mathbf{p}_i+\mathbf{t}. \label{eq:upnp_relation}
\end{equation}
Let $\mathbf{M}\in \mathbb{R}^{11\times 11}$ denote a coefficient matrix that embeds all known variables (point measurements and ray offsets), $\mathbf{q} = [q_0, q_1, q_2, q_3]^\top$ denote the quaternion representation of $\mathbf{R}$, and $\mathbf{s}\in \mathbb{R}^{11}$ denote a vector whose elements are all second and zeroth order monomials of the quaternion variables.
After translation and depth elimination, $\mathbf{q}$ can be constrained by first-order optimality conditions appearing in the form of the polynomial system\footnote{We kindly refer the reader to the UPnP paper for detailed derivations~\cite{kneip2014upnp}.}
\begin{equation}
    \begin{cases}
        \mathbf{s}^\top \, \mathbf{M} \, \frac{\partial \mathbf{s}}{\partial q_i} &= 0, \quad i = 0,..., 3\\
        \mathbf{q}^\top \mathbf{q} - 1 &= 0.
    \end{cases}\label{eq:UPnP_polys}
    \vspace{-0.5cm}
\end{equation}

\begin{figure}[t]
\begin{algorithm}[H]
\footnotesize 
\caption{Online Simulator HC for generalized PnP}
\label{alg:HCNet}
\begin{algorithmic}[1]
\State \textbf{Input}: correspondences $(\mathbf{p}_i, \mathbf{f}_i, \mathbf{v}_i)_{i = 1}^M$, stepsize $\delta t$
\State \textbf{Output}: poses $\hat{\mathbf{R}},\hat{\mathbf{t}}$
\State {$\mathring{\mathbf{R}},\mathring{\mathbf{t}} \leftarrow$ Trained Regression Network$((\mathbf{p}_i, \mathbf{f}_i, \mathbf{v}_i)_{i = 1}^M)$}
\State $(\hat{\mathbf{f}}_i)_{i = 1}^M  \leftarrow$ Reprojection  $(\mathring{\mathbf{R}},\mathring{\mathbf{t}};(\mathbf{p}_i, \mathbf{v}_i)_{i = 1}^M)$ 
\Comment{Online Simulator}
\State Starting problem $\mathbf{G}(\mathbf{x}) \leftarrow (\mathbf{p}_i, \hat{\mathbf{f}}_i, \mathbf{v}_i)_{i = 1}^M$ with solution $\mathring{\mathbf{R}},\mathring{\mathbf{t}}$ 
\State Target problem $\mathbf{F}(\mathbf{x}) \leftarrow (\mathbf{p}_i, \mathbf{f}_i, \mathbf{v}_i)_{i = 1}^M$
\State $\mathbf{q}_{HC} \leftarrow$ Homotopy Continuation $t\, \mathbf{F}(\mathbf{x}) + (1-t)\, \mathbf{G}(\mathbf{x})$
\State $\hat{\mathbf{R}},\hat{\mathbf{t}} \leftarrow$ decompose $\mathbf{q}_{HC}$
\end{algorithmic}
\end{algorithm}
\vspace{-0.6cm}
\end{figure}

\subsection{Regressor and online simulator}
\label{sec:upnp_HC}
To provide a proper problem-solution pair for HC, we train a pose regression network from synthetic data. Its input is given by row-wise stacking of the correspondences, i.e.
\begin{equation}
    \begin{bmatrix}
        \mathbf{p}_1^\top & \mathbf{f}_1^\top & \mathbf{v}_1^\top \\
        \vdots & \vdots & \vdots \\
        \mathbf{p}_N^\top & \mathbf{f}_N^\top & \mathbf{v}_N^\top 
    \end{bmatrix},
\end{equation}
where $N$ is the number of correspondences used, and the output is the quaternion vector. 
The network consists of $4$ 1D convolutional layers and a fully connected layer at the end to convert the output to a quaternion vector with the unit norm.
Given the rotation predicted by the network, the translation can be recovered from correspondences and estimated rotation. We refer the reader to \cite{kneip2014upnp} for the details.
For the online simulator, we can get reprojected $\hat{\mathbf{f}}_i$ from \eqref{eq:upnp_relation}.
Algorithm~\ref{alg:HCNet} summarizes all stages, including the final HC over \eqref{eq:UPnP_polys}. 
Note that in line 5 of Algorithm~\ref{alg:HCNet}, the starting problem $\mathbf{G}(\mathbf{x})$ is constructed by replacing the coefficient matrix $\mathbf{M}$ computed by $(\mathbf{p}_i, \mathbf{f}_i, \mathbf{v}_i)$ in \eqref{eq:UPnP_polys} with $\hat{\mathbf{M}}$ computed by $(\mathbf{p}_i, \hat{\mathbf{f}}_i, \mathbf{v}_i)$.

\subsection{Experiments}
\label{sec:upnp_experiments}

\noindent \textbf{Data.} We randomly generate $4$ cameras in a cube $[-1,1]^3$ and randomly sample $16$ 3D points in the box $[-1, 1]^2\times [4, 8]$. 16 samples would not be a good choice for a RANSAC solver \cite{fischler1981random}, but in the present context the idea is to set a moderate and notably equal challenge to all compared solvers. Furthermore, for UPnP the number of constraints appearing in the polynomial system remains unchanged as it analyses first-order optimality rather than individual incidence relations. Elements of the rotation angles are randomly generated in the interval $[-\pi/2, \pi/2]$, respectively, and the translation is randomly generated in the unit cube. We then generate the correspondences from rays that point to a 3D point.
The noise level is set to $2$ pixels.
To train the pose regression network, we simulate a dataset of $2400$ samples, which is sufficient for reliable Simulator HC initialization.
We use a 2.70GHz Intel(R) Xeon(R) CPU and an NVIDIA GeForce RTX 4090 for network training and evaluation.

\noindent \textbf{Methods.} For HC, the stepsize $\delta t$ and the maximum number of iterations of Newton's method are set to $0.02$ and $5$, respectively. We compare the Gr\"{o}bner Bases solver~\cite{kneip2014upnp} and the popular local method Levenberg Marquardt (LM)~\cite{ceres-solver}. 
We also test the state-of-the-art parallel HC solver developed by \cite{chien2022parallel, chien2022gpu}, and use both its CPU
and GPU\footnote{Note that GPU warmup overhead is ignored when averaging over multiple runs.} version. 
Given that parallel HC tracks all solutions, we manually pick the one with the smallest error. 

\noindent \textbf{Results.} The rotation error is measured by $\mathcal{E}_\mathbf{R} = 2\, \arcsin(\|\hat{\mathbf{R}}-\mathbf{R}_{gt}\|/(2\sqrt{2}))$ in degrees as suggested by \cite{ding2023minimal} and the translation error is measured by the $\mathcal{E}_\mathbf{t} = \|\hat{\mathbf{t}}-\mathbf{t}_{gt}\|/\|\mathbf{t}_{gt}\|\times 100$ in percentage. 
\tableautorefname~\ref{table:upnp2px} compares the traditional method, parallel exhaustive tracking HC, and different types of initial solutions for our simulator HC.
We present the success rate as the percentage of rotation errors smaller than $2$ degrees. It can be observed that on this simple task, except for the local method, traditional methods, as well as simulator HC using predicted initials, succeed in $100\%$ of trials. Meanwhile, online simulator HC benefits from the single-track strategy and turns out to be the fastest one. 
We also perform an ablation study on the proposed simulator HC, using both predicted and random initializations. Although the pose learned by the regressor is not highly accurate and only slightly outperforms random poses, it serves as a reasonable starting solution for our simulator HC. In contrast, LM as a local method is unable to leverage the potential of the predicted initializations, as the simulator HC does.

\begin{table}[t]
\centering
\caption{Performance comparison of simulator HC, a traditional method, and parallel exhaustive tracking HC on noisy synthetic data. All results are averaged over $1000$ trials. The success rate is computed for rotation errors with a threshold of $2$ degree.}
\label{table:upnp2px}
{\footnotesize
\setlength{\tabcolsep}{3pt}
\begin{tabular}{l|rcrc} 
\hline \hline
Method & Succ. (\%) & $\mathcal{E}_{\mathbf{R}}(deg)$& $\mathcal{E}_{\mathbf{t}}(\%)$ & Time(ms)  \\ 
\hline
Gr\"{o}bner Bases\; & 100 & 0.031 & 0.489 & 1.254  \\
CPU HC \; & 100  & 0.032& 0.499 & 9.796 \\
GPU HC \; & 100  & 0.032& 0.499 & 1.002  \\
\midrule
Pred Initial \; & 32 & 7.204& 32 & 0.031 \\
Pred Initial + LM \: & 54 & 28.72& 2256& 0.528\\
Rand Initial + Simulator HC \; & 22 & 91.68 & 228 & 0.231 \\
Pred Initial + Simulator HC \; & 100 & 0.032 & 0.513 & 0.316  \\ 
\hline \hline
\end{tabular}
}
\vspace{-0.5cm}
\end{table}

\section{Generalized Relative Pose and Scale}
\label{sec:GRPS}
Our final application targets a hard minimal problem: Generalized Relative Pose and Scale (GRPS) estimation from 2D-2D correspondences \cite{kneip2016generalized}.
The GRPS problem is a natural extension of the generalized relative pose problem and arises in point-less registration of partial view-graphs in structure-from-motion. Owing to the fact that SfM results are generally up to scale, an inherent scale factor needs to be resolved. 
Such problem was first introduced by Kneip et al.~\cite{kneip2016generalized}. It has 140 solutions if a unique rotation representation is used, and can be solved by multi-start optimization to overcome local minima.
Later on, a traditional polynomial solver based on the Gr\"{o}bner Bases method has been proposed~\cite{larsson2017efficient, martyushev2022optimizing}. With a $144\times 284$ elimination template, the solver finds all $140$ rotations within reasonable time.
In this section, we solve GRPS by our proposed method and compare it against the existing approach, as well as a parallel HC implementation tracking all roots.

\subsection{Problem Formulation}
Consider a 2D-2D generalized correspondence $(\mathbf{f}_i, \mathbf{v}_i)-(\mathbf{f}_i', \mathbf{v}_i')$ observed from two non-central camera views as illustrated in \figureautorefname~\ref{sub-fig:grps}. $\mathbf{f}_i\in\mathbb{S}^2, \mathbf{v}_i \in \mathbb{R}^3$ denote the image ray and ray origin in the first view, and $\mathbf{f}_i'\in\mathbb{S}^2, \mathbf{v}_i' \in \mathbb{R}^3$ are the corresponding entities in the second view.
The correspondence follows an underlying generalized relative pose and scale incidence relation depending on the rotation $\mathbf{R}\in SO(3)$, the position $\mathbf{t} \in \mathbb{R}^3$, the scale $s$, and unknown latent depths $\alpha_i$ and $\alpha_i'$ as follows:
\begin{equation}
    \alpha_i\mathbf{f}_i + \mathbf{v}_i = \mathbf{R}(\alpha_i' \mathbf{f}_i'+s \; \mathbf{v}_i') + \mathbf{t}. \label{eq:GRPS_relation}
\end{equation}
With the help of Pl\"{u}cker coordinates and the scalar triple product
\footnote{We kindly refer the reader to \cite{kneip2014efficient, kneip2016generalized} for detailed derivations}, the unknown depths get eliminated which results in a polynomial constraint that only involves the unknown poses
\begin{equation}
    \mathbf{t}^\top (\mathbf{f}_i \times \mathbf{R} \mathbf{f}_i')- s \mathbf{f}_i^\top \mathbf{R} [\mathbf{v}_i']_\times \mathbf{f}_i' + 
    \mathbf{f}_i^\top [\mathbf{v}_i]_\times \mathbf{R}\mathbf{f}_i' = 0.
    \label{eq:GRPS}
\end{equation}
Therefore, a minimal formulation for the GRPS problem can be devised by a polynomial system formed by stacking $7$ constraints in the form \eqref{eq:GRPS}, each one depending on one correspondence.



\subsection{Regressor and online simulator}
As introduced in \sectionautorefname~\ref{sec:upnp_HC}, we continue to use a pose regression network that inputs a matrix with row-wise stacked correspondences.
The output is given by $\mathbf{R}, \mathbf{t}, s$. Note that the rotation is predicted in continuous 6D representation which is more suitable for neural networks~\cite{zhou2019continuity}.
The network consists of $11$ 1D convolutional layers, in the end, it branches to $3$ sub-branches for $\mathbf{R}$, $\mathbf{t}$ and $s$. Each sub-branch consists of $2$ 1D convolutional layers and $1$ fully connected layers for adjust output.
Given the poses predicted by the network, we still need an estimation of the depths before simulating the reprojected image ray. 
With estimated $\hat{\mathbf{R}}, \hat{\mathbf{t}}$ and $\hat{s}$, the corresponding depths can be obtained from \eqref{eq:GRPS_relation} by following
\begin{equation}
    \begin{bmatrix}
        \mathbf{f}_i &
        -\hat{\mathbf{R}} \mathbf{f}_i'
    \end{bmatrix} 
    \begin{bmatrix}
        \alpha_i \\
        \alpha_i'
    \end{bmatrix}= 
    -(\mathbf{v}_i - \hat{s} \hat{\mathbf{R}} \mathbf{v}_i' - \hat{\mathbf{t}}),
    \label{eq:grps_getdepths}
\end{equation}
which can be simply solved by linear least squares for each input correspondence.
Finally, we simulate a pose-consistent reprojected image ray in the first view using
\begin{equation}
    \hat{\mathbf{f}}_i = \frac{1}{\hat{\alpha_i}}\hat{\mathbf{R}}(\hat{\alpha_i}' \mathbf{f}_i'+\hat{s} \; \mathbf{v}_i') + \hat{\mathbf{t}} - \mathbf{v}_i.
\end{equation}


\subsection{Simulation Experiments}

\noindent \textbf{Data.}
We adopt a similar data generation method as in \sectionautorefname~\ref{sec:upnp_experiments}. We generate $7$ 3D points in a $[-1,1]^2\times [2, 20]$ volume. Two view frames are generated with origins picked from $[-1,1]^3$. Within each view frame, we then sample individual camera offsets from the volume $[-1,1]^3$. The scale is randomly generated in $[0.1,5.0]$, and all other parameters related to noise addition are left unchanged. Note that the ground truth relative pose needs to be extracted from the simulated pair of view poses.

\noindent \textbf{Methods.}
The stepsize for homotopy continuation is set to $0.05$ in this scenario. We compare against the Eigen solver~\cite{kneip2016generalized}, Gr\"obner Bases solver~\cite{martyushev2022optimizing} and the CPU and GPU versions of parallel HC which tracks $140$ solutions found by Monodromy~\cite{breiding18}.
Note that to measure numerical errors, we just pick the best solution with the smallest error. In practice, traditional HC needs to evaluate additional information to select a single solution (\eg euclidean distance). For a complicated problem like GRPS, evaluating $140$ solutions would still induce non-neglectable time-consumption.
Since GRPS is a hard problem, we train the regressor using a larger simulated dataset of size $64000$.

\noindent \textbf{Metric.} We measure the rotation error in degrees and the relative translation as in \cref{sec:upnp_experiments} and scale error in percentage using
\begin{equation}
\mathcal{E}_s = | \hat{s} - s_{gt} | / s_{gt}  \times 100.
\end{equation}
To calculate the success rate, for rotation errors the threshold is set to $2$ deg, and for relative errors, it is set to $5\%$.


\noindent \textbf{Numerical Stability.}
We randomly generate $1000$ trials to test the numerical stability of all solvers.
During the analysis of the performance of simulator HC using the minimum number of correspondences, we observed that our method has around $70\%$ success rate. By checking the failure cases, we found that sometimes simulator HC tracked another real solution that was different from the ground truth pose. 
To tackle that issue, we evaluate the addition of one more correspondence and solve for an overdetermined polynomial system. By using $8$ correspondences, simulator HC reaches a success rate of $95\%$. 
Again, since CPU/GPU-HC is designed to handle square systems, only the minimum number of correspondences can be tested for other HC methods.

Since the errors in rotation, translation and scale have similar density results, we only present the rotation error in the main paper and the rest in the supplementary. As illustrated in \figureautorefname~\ref{fig:numerical_stability_R}, the minimal HC solvers all present similar error distribution and around $75\%$ of trials are successful. While Gr\"obner Bases shows $100\%$ success rate, Simulator HC over 8pts presents similar numerical errors.

\begin{figure*}[t!]
    \centering
    \begin{subfigure}[t]{0.32\textwidth}
        \centering
        \includegraphics[width=\linewidth]{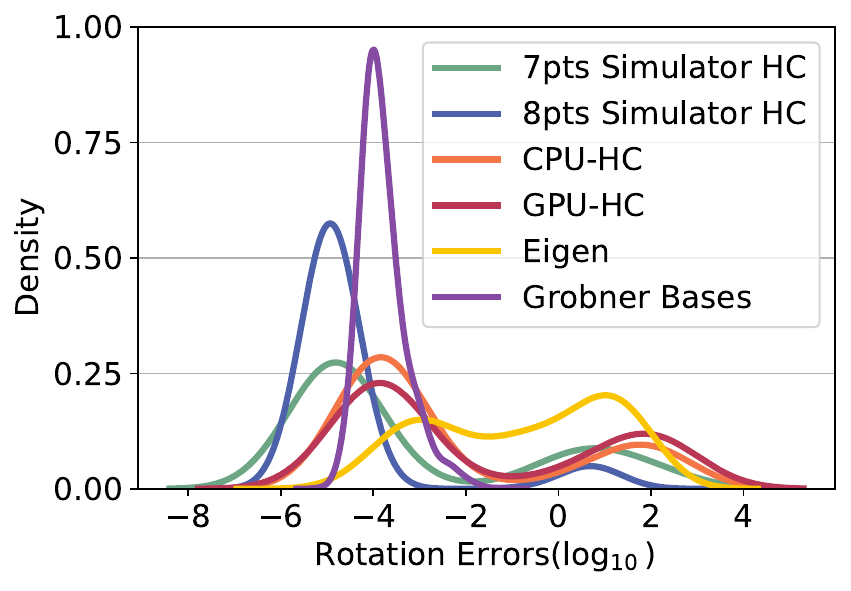}
        \caption{Error distribution over noise-free data.}
        \label{fig:numerical_stability_R}
    \end{subfigure}%
    \hfill
    \begin{subfigure}[t]{0.34\textwidth}
        \centering
        \includegraphics[width=\linewidth]{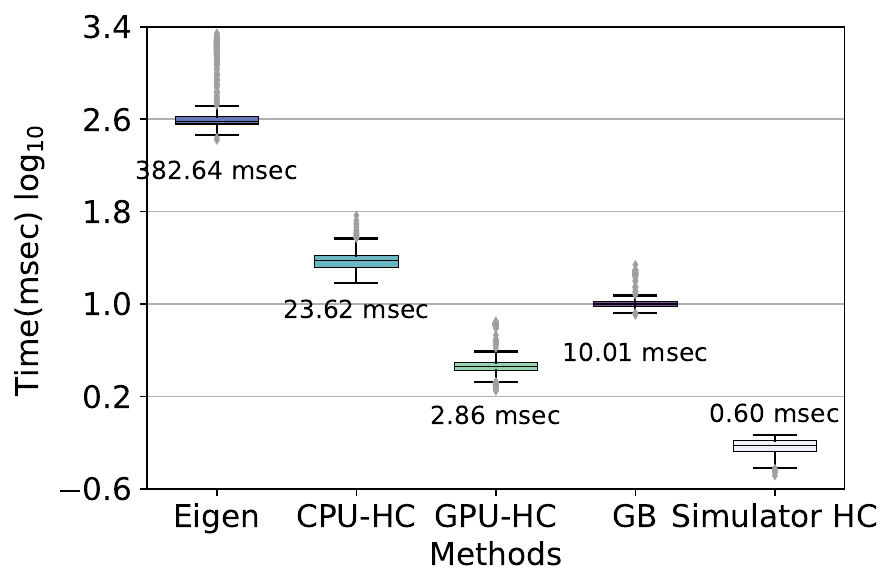}
        \caption{The boxplot of running time comparison. }
        \label{fig:time_box}
    \end{subfigure}%
    \hfill
    \begin{subfigure}[t]{0.32\textwidth}
        \centering
        \includegraphics[width=\linewidth]{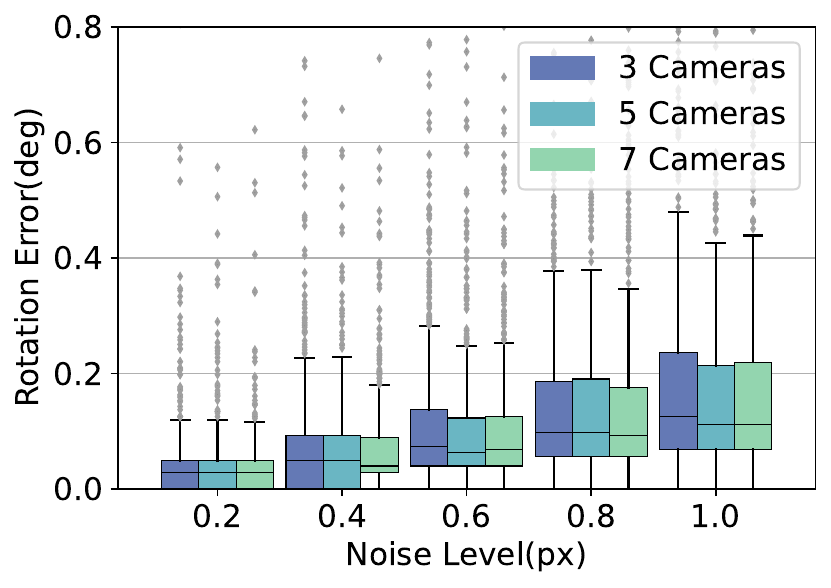}
        \caption{Error statistics of simulator HC.}
        \label{fig:GRPS_noisy}
    \end{subfigure}
    \caption{ Experimental results on GRPS. \cref{fig:numerical_stability_R} Error distribution over $1000$ trials on noise-free data. The camera number is set to be $3$. Considering the rotation error, except for Gr\"obner Bases, the other minimal solvers all present a similar error distribution with around $70\%$ success rate, and the eigen solver presents only around $60\%$ success rate. The proposed $8$pts simulator HC presents $96.3\%$ success rate where the Gr\"obner Bases has $100\%$ success rate. \cref{fig:time_box} The boxplot of running time comparison for the proposed simulator HC and other methods, where the number represents the median time. As expected, on average the proposed online simulator HC running on CPU is the fastest. It is about $5\times$ faster than GPU-HC since we only track one solution, $17\times$ faster than Gr\"obner Bases and about $40\times$ faster than CPU-HC. \cref{fig:GRPS_noisy} Error statistics of simulator HC with respect to different noise levels ranging from $0.2$ pixel to $1.0$ pixel.}
    \label{fig:combined_figures}
\vspace{-0.4cm}
\end{figure*}




\noindent \textbf{Time Comparison.}
The main computational benefit of the proposed online simulator HC comes from single solution-tracking, which can be effectuated efficiently even on the CPU. \figureautorefname~\ref{fig:time_box} demonstrates the efficiency gain of our $8$pts simulator HC over the eigen solver and parallel HC variants. As expected, by tracking only one solution curve, we are significantly faster than parallel exhaustive tracking HC, even if the latter runs on GPU. The eigen solver is an optimization-based solver with high complexity and ends up being slowest. The Gr\"obner Bases solver dealing with the large elimination template still costs a considerable time, even for just computing solutions for rotation, and not translation or scale.

\begin{remark}
\vspace{-0.2cm}
    We would like to note that Gr\"obner Bases is solving a different formulation from \cref{eq:GRPS}. In order to make it easy to be solved by a smaller elimination template, it eliminates translation and scale~\cite{stewenius05}. But the resulting coefficient matrix is $130\times 275$ and still needs a large number of operations. Although the formulation of \cref{eq:GRPS} with eliminated depths faces a slightly lower success rate, its number of operations is much smaller than Gr\"obner Bases. On the other hand, the proposed 8pt simulator HC provides a comparably high success rate as Gr\"obner Bases and is still $17\times$ faster. 
\end{remark}


\begin{table}[ht]
\vspace{-0.3cm}
\centering
\caption{RANSAC experiments for the GRPS problem using Gr\"{o}bner Bases $\mid$ simulator HC. Image points are randomly perturbed by 2-pixel uniform noise. We use vanilla RANSAC with at most $200$ iterations. As can be observed, outlier ratios of up to $40\%$ are easily handled within 200 iterations. And regressor-based simulator HC is consistently faster than Gr\"{o}bner Bases.} 
\label{table:outlier}
{\footnotesize
\setlength{\tabcolsep}{5pt}
\begin{tabular}{c|cccc}
\hline \hline
\text{Ratio} & \text{Succ. (\%)} & $\mathcal{E}_{\mathbf{R}}$(\text{deg}) & \# \text{ Iters} & \text{Time(s)}\\ 
\hline
10\% & \;\,99.5 $\mid$ 100 & 0.10 $\mid$ 0.09 & \;\,93 $\mid$ 125 & 2.33 $\mid$ 0.85\\
20\% & \;\,99.4 $\mid$ 100 & 0.11 $\mid$ 0.10 & 171 $\mid$ 193 & 4.37 $\mid$ 1.31\\
30\% & \;\,99.4 $\mid$ 100 & 0.13 $\mid$ 0.14 & 200 $\mid$ 200 & 5.10 $\mid$ 1.35 \\
40\% & 98.9 $\mid$ 91 & 0.21 $\mid$ 0.29 & 200 $\mid$ 200 & 4.59 $\mid$ 1.35\\
\hline \hline
\end{tabular}
}
\end{table}

\noindent \textbf{Noise Resilience.}
\figureautorefname~\ref{fig:GRPS_noisy} presents the performance of the proposed $8$pts simulator HC with different numbers of cameras under increasing image noise levels ranging from $0.2$ pixel to $1.0$ pixel. The error behaves gracefully as a function of noise, and it can be observed that a higher number of cameras leads to slightly inferior noise. Furthermore, we also observe that rotation errors appear to be less sensitive to image noise than translation and scale errors.\footnote{shown in supplementary.}


\noindent \textbf{Evaluation inside RANSAC.}
To conclude, we embed simulator HC into RANSAC to test its practical usefulness.
We synthesize $200$ correspondences between $5$ cameras and replace part of them with outliers.
Uniformly sampled pixel noise from $[-2,2]$ is added to both coordinates of the image points by again assuming a virtual focal length of 800 px.
We terminate RANSAC if it reaches $99\%$ confidence or $200$ iterations, and the inlier threshold is set to $0.01$ for the distance measured by the normalized reprojection error. 
Given that Gr\"{o}bner Bases solver estimates $140$ solutions, we keep the real ones and select the best solution using reprojection errors in parallel to save time for it.
\tableautorefname~\ref{table:outlier} shows the results. 
The success rate is measured by a rotation error threshold of $2$ degrees and the indicated errors are the average accuracy of successful cases, results on translation and scale are left in supplementary. Overall, simulator HC embedded into RANSAC can easily handle up to $40\%$ outliers as Gr\"{o}bner Bases and much faster, which suggests high practicality. Note that Gr\"{o}bner Bases provides around 46 real solutions each time, which requires a significant amount of additional time for solution selection. Meanwhile, simulator HC does not require this extra step.

\begin{figure}[b]
    \centering
    \includegraphics[width=\linewidth]{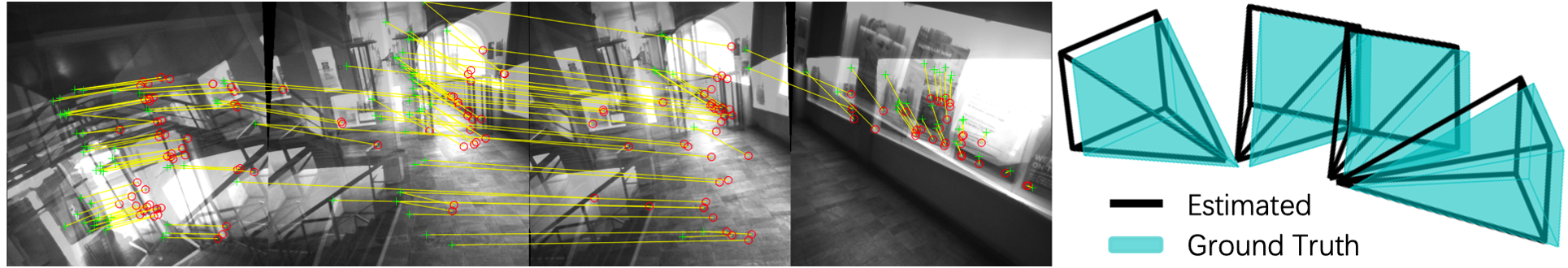}
    \caption{LaMAR dataset multi-camera rig visualization.}
    \label{fig:HoloLens}
\end{figure}

\begin{table}[b]
    \centering
    \caption{LaMAR dataset RANSAC-simulatorHC Median results.}
    \label{tab:LaMAR}
    {\footnotesize
    \setlength{\tabcolsep}{3pt}
    \begin{tabular}{c|c|c|c|c|c|c|c}
    \hline \hline
     & seq. & $\mathcal{E}_\mathbf{R}$(deg) & $\mathcal{E}_s(\%)$ & $\mathcal{E}_{\mathbf{t}}$(deg) & \#iters & Time(s) & Succ. (\%) \\ \hline
    \multirow{3}{*}{\makecell{Gr\"{o}bner \\ Bases}} & CAB & 1.18 & 56.14 & \textbf{5.70} & \textbf{137} & 2.83 & 60 \\
    & HGE & 0.62 & 62.48 & \textbf{4.39} & \textbf{59} & 1.25 & 67 \\
    & LIN & 0.60 & 42.04 & \textbf{3.40} & \textbf{31} & 0.65 & 72 \\ 
    \midrule
    \multirow{3}{*}{\makecell{simulator \\ HC}} & CAB & \textbf{1.05} & \textbf{43.20} & 9.16 & 310 & \textbf{2.08} & \textbf{68} \\
    & HGE & \textbf{0.56} & \textbf{41.52} & 7.19 & 134 & \textbf{0.90} & \textbf{84} \\
    & LIN & \textbf{0.56} & \textbf{32.19} & 4.32 & 66 & \textbf{0.44} & \textbf{86} \\ 
    \hline \hline
    \end{tabular}
    }
\end{table}

\subsection{Real World Experiments}
We further explore the performance of simulator HC on the real dataset LaMAR~\cite{sarlin2022lamar}.
As illustrated in~\cref{fig:HoloLens}, the provided HoloLens data is a natural application of GRPS with identity scale, which can be viewed as a multi-camera rig with $4$ cameras in the front and sides. We extract the correspondences by consecutively taking a view-pair from the trajectories and ensuring the distance between each view is in $[0.2m, 0.5m]$ with at least $100$ matches. The resulting number of pairs is $1794$ (CAB), $2852$ (HGE) and $739$ (LIN) in total. The correspondences are matched with SIFT features using nearest neighbour matching.
We also present the angular error in translation, as in practice, the scale of translation and overall scale could be unobservable, \eg the degenerate scenario when correspondences remain on the frame captured by the same camera.\\

\Cref{tab:LaMAR} shows the median results over all three sequences of LaMAR dataset. We adopted a similar setting for RANSAC as the above section with $500$ maximum iterations. The proposed solver still presents consistent performances, where simulator HC has higher success rate with faster running time than Gr\"{o}bner Bases.

\section{Conclusion}
\label{sec: conclusion}

The present paper introduces a new paradigm for correspondence-based geometric problem solving using an elegant three-stage combination of a regression network, an online correspondence simulator, and homotopy continuation. The regression network acts as a general solution approximator and can be trained in simulation only. While pure regression accuracy is often insufficient on its own, a key message conveyed through our results is that the accuracy is generally still good enough to simulate a consistent starting problem-solution pair that successfully enables single-solution continuation. As demonstrated on two generalized camera problems, the strategy enables high success rates on common computing hardware, and the resulting solvers prove to be a viable solution if utilized as part of a random sample and consensus scheme. In particular, we propose a successful solution to the generalized relative pose and scale problem, which is around $17\times$ faster than traditional polynomial elimination techniques.

\section{Acknowledgments}
The authors would like to acknowledge the funding support provided by projects 22DZ1201900, 22ZR1441300, and DFYJBJ-1 by the Natural Science Foundation of Shanghai.

{
    \small
    \bibliographystyle{ieeenat_fullname}
    \bibliography{main}

\begin{thebibliography}{66}
\providecommand{\natexlab}[1]{#1}
\providecommand{\url}[1]{\texttt{#1}}
\expandafter\ifx\csname urlstyle\endcsname\relax
  \providecommand{\doi}[1]{doi: #1}\else
  \providecommand{\doi}{doi: \begingroup \urlstyle{rm}\Url}\fi

\bibitem[Agarwal et~al.(2022)Agarwal, Mierle, and Others]{ceres-solver}
Sameer Agarwal, Keir Mierle, and Others.
\newblock Ceres solver.
\newblock \url{http://ceres-solver.org}, 2022.

\bibitem[Aholt and Oeding(2014)]{aholt14}
C. Aholt and L. Oeding.
\newblock The ideal of the trifocal variety.
\newblock \emph{Math. Comput.}, 83\penalty0 (289):\penalty0 2553–--2574, 2014.

\bibitem[Amos and Kolter(2017)]{amos2017optnet}
B. Amos and J.~Z. Kolter.
\newblock Optnet: Differentiable optimization as a layer in neural networks.
\newblock In \emph{International Conference on Machine Learning}, pages 136--145. PMLR, 2017.

\bibitem[Bates et~al.(2013)Bates, Sommese, Hauenstein, and Wampler]{bates13}
D.~J. Bates, A.~J. Sommese, J.~D. Hauenstein, and C.~W. Wampler.
\newblock Numerically solving polynomial systems with {Bertini}.
\newblock \emph{Software, environments, tools, SIAM}, 25, 2013.

\bibitem[Bhayani et~al.(2020)Bhayani, Kukelova, and Heikkil{\"a}]{bhayani20}
S. Bhayani, Z. Kukelova, and J. Heikkil{\"a}.
\newblock A sparse resultant based method for efficient minimal solvers.
\newblock In \emph{CVPR}, 2020.

\bibitem[Breiding and Timme(2018)]{breiding18}
P. Breiding and S. Timme.
\newblock Homotopycontinuation.jl: A package for homotopy continuation in {Julia}.
\newblock In \emph{International Conference on Mathematical Software - ICMS}, 2018.

\bibitem[Chien et~al.(2022{\natexlab{a}})Chien, Fan, Abdelfattah, Tsigaridas, Tomov, and Kimia]{chien2022gpu}
C.-H. Chien, H. Fan, A. Abdelfattah, E. Tsigaridas, S. Tomov, and B. Kimia.
\newblock Gpu-based homotopy continuation for minimal problems in computer vision.
\newblock In \emph{CVPR}, 2022{\natexlab{a}}.

\bibitem[Chien et~al.(2022{\natexlab{b}})Chien, Fan, Abdelfattah, Tsigaridas, Tomov, and Kimia]{chien2022parallel}
C.-H. Chien, H. Fan, A. Abdelfattah, E. Tsigaridas, S. Tomov, and B. Kimia.
\newblock Parallel path tracking for homotopy continuation using gpu.
\newblock In \emph{Proceedings of the International Symposium on Symbolic and Algebraic Computation}, 2022{\natexlab{b}}.

\bibitem[Cox et~al.(1998)Cox, Little, and O’Shea]{cox98}
D. Cox, J. Little, and D. O’Shea.
\newblock \emph{Using Algebraic Geometry}.
\newblock Springer, 1998.

\bibitem[Dai et~al.(2016)Dai, Li, and Kneip]{dai16}
Y. Dai, H. Li, and L. Kneip.
\newblock Rolling shutter camera relative pose: Generalized epipolar geometry.
\newblock In \emph{CVPR}, 2016.

\bibitem[Ding et~al.(2023)Ding, Chien, Larsson, \r{A}str\"om, and Kimia]{ding2023minimal}
Y. Ding, C.-H. Chien, V. Larsson, K. \r{A}str\"om, and B. Kimia.
\newblock Minimal solutions to generalized three-view relative pose problem.
\newblock In \emph{ICCV}, 2023.

\bibitem[Duff et~al.(2018)Duff, Hill, Jensen, Lee, Leykin, and Sommars]{duff18}
T. Duff, C. Hill, A. Jensen, K. Lee, A. Leykin, and J. Sommars.
\newblock Solving polynomial systems via homotopy continuation and monodromy.
\newblock \emph{IMA Journal of Numerical Analysis}, 2018.

\bibitem[Duff et~al.(2019)Duff, Kohn, Leykin, and Pajdla]{duff19}
T. Duff, K. Kohn, A. Leykin, and T. Pajdla.
\newblock {PLMP} - point-line minimal problems in complete multi-view visibility.
\newblock In \emph{ICCV}, pages 1675--1684, 2019.

\bibitem[Duff et~al.(2020)Duff, Kohn, Leykin, and Pajdla]{duff20}
T. Duff, K. Kohn, A. Leykin, and T. Pajdla.
\newblock {PL1P} - point-line minimal problems under partial visibility in three views.
\newblock In \emph{ECCV}, 2020.

\bibitem[Elqursh and Elgammal(2011)]{elqursh11}
A. Elqursh and A.~M. Elgammal.
\newblock Line-based relative pose estimation.
\newblock In \emph{CVPR}, 2011.

\bibitem[Emiris(2012)]{emiris2012general}
I.~Z. Emiris.
\newblock A general solver based on sparse resultants, 2012.

\bibitem[Fabbri et~al.(2020)Fabbri, Duff, Fan, Regan, da~Costa~de Pinho, Tsigaridas, Wampler, Hauenstein, Giblin, Kimia, and Pajdla]{fabbri20}
R. Fabbri, T. Duff, H. Fan, M.~H. Regan, D. da~Costa~de Pinho, E.~P. Tsigaridas, C.~W. Wampler, J.~D. Hauenstein, P.~J. Giblin, A. Kimia, B.~B.~Leykin, and T. Pajdla.
\newblock {TR-PLP - Trifocal relative pose from lines at points}.
\newblock In \emph{CVPR}, 2020.

\bibitem[Fischler and Bolles(1981)]{fischler1981random}
Martin~A Fischler and Robert~C Bolles.
\newblock Random sample consensus: a paradigm for model fitting with applications to image analysis and automated cartography.
\newblock \emph{Communications of the ACM}, 24\penalty0 (6):\penalty0 381--395, 1981.

\bibitem[Fischler and Bolles(1985)]{fischler81}
M.~A. Fischler and R.~C. Bolles.
\newblock Random sample consensus: a paradigm for model fitting with applications to image analysis and automated cartography.
\newblock \emph{Commun. ACM}, 24\penalty0 (6):\penalty0 381--395, 1985.

\bibitem[Haralick et~al.(1994)Haralick, Lee, Ottenberg, and N\"olle]{haralick94}
R.~M. Haralick, C.-N. Lee, K. Ottenberg, and M. N\"olle.
\newblock Review and analysis of solutions of the three point perspective pose estimation problem.
\newblock \emph{IJCV}, 13\penalty0 (3):\penalty0 331–--356, 1994.

\bibitem[Hartley and Zisserman(2004)]{hartley2004multiple}
R. Hartley and A. Zisserman.
\newblock \emph{Multiple View Geometry in Computer Vision}.
\newblock Cambridge University Press, 2nd edition, 2004.

\bibitem[Heikkil{\"a}(2017)]{heikkila17}
J. Heikkil{\"a}.
\newblock Using sparse elimination for solving minimal problems in computer vision.
\newblock In \emph{ICCV}, 2017.

\bibitem[Hesch and Roumeliotis(2011)]{hesch11}
J. Hesch and S.~I. Roumeliotis.
\newblock {A Direct Least-Squares (DLS) method for PnP}.
\newblock In \emph{ICCV}, 2011.

\bibitem[Holt and Netravali(1995)]{holt95}
R.~J. Holt and A.~N. Netravali.
\newblock Uniqueness of solutions to three perspective views of four points.
\newblock \emph{IEEE TPAMI}, 17\penalty0 (3):\penalty0 303–--307, 1995.

\bibitem[Hruby et~al.(2022)Hruby, Duff, Leykin, and Pajdla]{hruby2022learning}
P. Hruby, T. Duff, A. Leykin, and T. Pajdla.
\newblock Learning to solve hard minimal problems.
\newblock In \emph{CVPR}, 2022.

\bibitem[Kendall and Cipolla(2017)]{kendall2017geometric}
A. Kendall and R. Cipolla.
\newblock Geometric loss functions for camera pose regression with deep learning.
\newblock In \emph{CVPR}, pages 5974--5983, 2017.

\bibitem[Kileel(2017)]{kileel17}
J. Kileel.
\newblock Minimal problems for the calibrated trifocal variety.
\newblock \emph{SIAM Journal on Applied Algebra and Geometry}, 1\penalty0 (1):\penalty0 575–--598, 2017.

\bibitem[Kneip and Li(2014)]{kneip2014efficient}
Laurent Kneip and Hongdong Li.
\newblock Efficient computation of relative pose for multi-camera systems.
\newblock In \emph{Proceedings of the IEEE conference on computer vision and pattern recognition}, pages 446--453, 2014.

\bibitem[Kneip et~al.(2011)Kneip, Scaramuzza, and Siegwart]{kneip11}
L. Kneip, D. Scaramuzza, and R. Siegwart.
\newblock A novel parametrization of the perspective-three-point problem for a direct computation of absolute camera position and orientation.
\newblock In \emph{CVPR}, 2011.

\bibitem[Kneip et~al.(2012)Kneip, Siegwart, and Pollefeys]{kneip12}
L. Kneip, R. Siegwart, and M. Pollefeys.
\newblock Finding the exact rotation between two images independently of the translation.
\newblock In \emph{ECCV}, 2012.

\bibitem[Kneip et~al.(2014)Kneip, Li, and Seo]{kneip2014upnp}
L. Kneip, H. Li, and Y. Seo.
\newblock {UPnP: An optimal O(n) solution to the absolute pose problem with universal applicability}.
\newblock In \emph{ECCV}, 2014.

\bibitem[Kneip et~al.(2016)Kneip, Sweeney, and Hartley]{kneip2016generalized}
L. Kneip, C. Sweeney, and R. Hartley.
\newblock The generalized relative pose and scale problem: View-graph fusion via 2d-2d registration.
\newblock 2016.

\bibitem[Kukelova et~al.(2008)Kukelova, Bujnak, and Pajdla]{kukelova08}
Z. Kukelova, M. Bujnak, and T. Pajdla.
\newblock Automatic generator of minimal problem solvers.
\newblock In \emph{ECCV}, 2008.

\bibitem[Larsson et~al.(2017{\natexlab{a}})Larsson, Astr{\"o}m, and Oskarsson]{larsson17a}
V. Larsson, K. Astr{\"o}m, and M. Oskarsson.
\newblock Efficient solvers for minimal problems by syzygy-based reduction.
\newblock In \emph{CVPR}, 2017{\natexlab{a}}.

\bibitem[Larsson et~al.(2017{\natexlab{b}})Larsson, Astr{\"o}m, and Oskarsson]{larsson17b}
V. Larsson, K. Astr{\"o}m, and M. Oskarsson.
\newblock Polynomial solvers for saturated ideals.
\newblock In \emph{ICCV}, 2017{\natexlab{b}}.

\bibitem[Larsson et~al.(2017{\natexlab{c}})Larsson, Astrom, and Oskarsson]{larsson2017efficient}
Viktor Larsson, Kalle Astrom, and Magnus Oskarsson.
\newblock Efficient solvers for minimal problems by syzygy-based reduction.
\newblock In \emph{Proceedings of the IEEE Conference on Computer Vision and Pattern Recognition}, pages 820--829, 2017{\natexlab{c}}.

\bibitem[Larsson et~al.(2017{\natexlab{d}})Larsson, Kukelova, and Zheng]{larsson17c}
V. Larsson, Z. Kukelova, and Y. Zheng.
\newblock Making minimal solvers for absolute pose estimation compact and robust.
\newblock In \emph{ICCV}, 2017{\natexlab{d}}.

\bibitem[Larsson et~al.(2018)Larsson, Oskarsson, Astr{\"o}m, Wallis, Kukelova, and Pajdla]{larsson18}
V. Larsson, M. Oskarsson, K. Astr{\"o}m, A. Wallis, Z. Kukelova, and T. Pajdla.
\newblock Beyond gr{\"o}bner bases: Basis selection for minimal solvers.
\newblock In \emph{CVPR}, 2018.

\bibitem[Leykin(2011)]{leykin11}
A. Leykin.
\newblock Numerical algebraic geometry.
\newblock \emph{Journal of Software for Algebra and Geometry}, 3\penalty0 (1):\penalty0 5--–10, 2011.

\bibitem[Martyushev et~al.(2022)Martyushev, Vrablikova, and Pajdla]{martyushev2022optimizing}
Evgeniy Martyushev, Jana Vrablikova, and Tomas Pajdla.
\newblock Optimizing elimination templates by greedy parameter search.
\newblock In \emph{Proceedings of the IEEE/CVF Conference on Computer Vision and Pattern Recognition}, pages 15754--15764, 2022.

\bibitem[Miraldo et~al.(2018)Miraldo, Dias, and Ramalingam]{miraldo18}
P. Miraldo, T. Dias, and S. Ramalingam.
\newblock A minimal closed-form solution for multi-perspective pose estimation using points and lines.
\newblock In \emph{ECCV}, 2018.

\bibitem[Monga et~al.(2021)Monga, Li, and Eldar]{monga2021algorithm}
V. Monga, Y. Li, and Y.~C. Eldar.
\newblock Algorithm unrolling: Interpretable, efficient deep learning for signal and image processing.
\newblock \emph{IEEE Signal Processing Magazine}, 38\penalty0 (2):\penalty0 18--44, 2021.

\bibitem[Mur-Artal et~al.(2015)Mur-Artal, Montiel, and D.]{murartal15}
R. Mur-Artal, J.~M.~M. Montiel, and Tard{\'o}s.~J. D.
\newblock {ORB-SLAM: A Versatile and Accurate Monocular SLAM System}.
\newblock 31\penalty0 (5), 2015.

\bibitem[Nist{\'e'}r(2004)]{nister2004efficient}
D. Nist{\'e'}r.
\newblock An efficient solution to the five-point relative pose problem.
\newblock \emph{IEEE TPAMI}, 26\penalty0 (6):\penalty0 756--770, 2004.

\bibitem[Nist{\'e'}r and Schaffalitzky(2006)]{nister06}
D. Nist{\'e'}r and F. Schaffalitzky.
\newblock Four points in two or three calibrated views: Theory and practice.
\newblock \emph{IJCV}, 67\penalty0 (2):\penalty0 211--231, 2006.

\bibitem[Nist{\'e}r et~al.(2004)Nist{\'e}r, Naroditsky, and Bergen]{nister04}
D. Nist{\'e}r, O. Naroditsky, and J. Bergen.
\newblock Visual odometry.
\newblock In \emph{CVPR}, 2004.

\bibitem[Quan et~al.(2006)Quan, Triggs, and Mourrain]{quan06}
L. Quan, B. Triggs, and B. Mourrain.
\newblock Some results on minimal euclidean reconstruction from four points.
\newblock 24\penalty0 (3):\penalty0 341–--348, 2006.

\bibitem[Raguram et~al.(2013)Raguram, Chum, Pollefeys, Matas, and Frahm]{raguram13}
R. Raguram, O. Chum, M. Pollefeys, J. Matas, and J.-M. Frahm.
\newblock {SAC: A universal framework for random sample consensus}.
\newblock \emph{IEEE TPAMI}, 35\penalty0 (8):\penalty0 2022–--2038, 2013.

\bibitem[Sarlin et~al.(2021)Sarlin, Unagar, Larsson, Germain, Toft, Larsson, Pollefeys, Lepetit, Hammarstrand, Kahl, et~al.]{sarlin2021back}
P.-E. Sarlin, A. Unagar, M. Larsson, H. Germain, C. Toft, V. Larsson, M. Pollefeys, V. Lepetit, L. Hammarstrand, F. Kahl, et~al.
\newblock Back to the feature: Learning robust camera localization from pixels to pose.
\newblock In \emph{CVPR}, 2021.

\bibitem[Sarlin et~al.(2022)Sarlin, Dusmanu, Schönberger, Speciale, Gruber, Larsson, Miksik, and Pollefeys]{sarlin2022lamar}
Paul-Edouard Sarlin, Mihai Dusmanu, Johannes~L. Schönberger, Pablo Speciale, Lukas Gruber, Viktor Larsson, Ondrej Miksik, and Marc Pollefeys.
\newblock {LaMAR: Benchmarking Localization and Mapping for Augmented Reality}.
\newblock In \emph{ECCV}, 2022.

\bibitem[Saurer et~al.(2015)Saurer, Pollefeys, and Lee]{saurer15}
O. Saurer, M. Pollefeys, and G.~H. Lee.
\newblock A minimal solution to the rolling shutter pose estimation problem.
\newblock 2015.

\bibitem[Sch{\"o}nberger and Frahm(2016)]{schonberger16}
J. Sch{\"o}nberger and J.-M. Frahm.
\newblock Structure-from-motion revisited.
\newblock In \emph{CVPR}, 2016.

\bibitem[Sheffer and Wiesel(2020)]{sheffer2020pnpnet}
R. Sheffer and A. Wiesel.
\newblock {PnP-Net: A hybrid Perspective-n-Point Network}, 2020.

\bibitem[Snavely et~al.(2006)Snavely, Seitz, and Szeliski]{snavely06}
N. Snavely, S.~M. Seitz, and R. Szeliski.
\newblock {Photo tourism: exploring photo collections in 3D}.
\newblock 2006.

\bibitem[Snavely et~al.(2008)Snavely, Seitz, and Szeliski]{snavely08}
N. Snavely, S.~M. Seitz, and R. Szeliski.
\newblock Modeling the world from internet photo collections.
\newblock pages 189--210, 2008.

\bibitem[Sommese et~al.(2005)Sommese, Wampler, et~al.]{sommese2005numerical}
Andrew~J Sommese, Charles~W Wampler, et~al.
\newblock \emph{The Numerical solution of systems of polynomials arising in engineering and science}.
\newblock World Scientific, 2005.

\bibitem[Stew{\'e}nius et~al.(2005)Stew{\'e}nius, Nist{\'e}r, Oskarsson, and Astr{\"o}m]{stewenius05}
H. Stew{\'e}nius, D. Nist{\'e}r, M. Oskarsson, and K. Astr{\"o}m.
\newblock Solutions to minimal generalized relative pose problems.
\newblock In \emph{Workshop on Omnidirectional Vision in conjunction with ICCV}, 2005.

\bibitem[Stewenius et~al.(2006)Stewenius, Engels, and Nist{\'e'}r]{stewenius06}
H. Stewenius, C. Engels, and D. Nist{\'e'}r.
\newblock Recent developments on direct relative orientation.
\newblock \emph{ISPRS J. of Photogrammetry and Remote Sensing}, 60:\penalty0 284--294, 2006.

\bibitem[Sturmfels(2002)]{sturmfels02}
B. Sturmfels.
\newblock Solving systems of polynomial equations, 2002.

\bibitem[Ventura et~al.(2015)Ventura, Arth, and Lepetit]{ventura15}
J. Ventura, C. Arth, and V. Lepetit.
\newblock An efficient minimal solution for multi-camera motion.
\newblock In \emph{ICCV}, 2015.

\bibitem[Verschelde(2010)]{verschelde10}
J. Verschelde.
\newblock Polynomial homotopy continuation with phcpack.
\newblock \emph{ACM Commun. Comput. Algebra}, 44\penalty0 (3/4):\penalty0 217–--220, 2010.

\bibitem[Xu et~al.(2019)Xu, Lan, Tsakiris, and Kneip]{xu19}
W. Xu, H. Lan, M.~C. Tsakiris, and L. Kneip.
\newblock Online stability improvement of gr\"obner basis solvers using deep learning.
\newblock 2019.

\bibitem[Zhao et~al.(2020)Zhao, Kneip, He, and Ma]{zhao20}
J. Zhao, L. Kneip, Y. He, and J. Ma.
\newblock Minimal case relative pose computation using ray-point-ray features.
\newblock \emph{IEEE TPAMI}, 42:\penalty0 1176--1190, 2020.

\bibitem[Zheng and Kneip(2016)]{zheng16}
Y. Zheng and L. Kneip.
\newblock {A direct least-squares solution to the PnP problem with unknown focal length}.
\newblock In \emph{CVPR}, 2016.

\bibitem[Zheng et~al.(2013)Zheng, Kuang, Sugimoto, Astrom, and Okutomi]{zheng13}
Y. Zheng, Y. Kuang, S. Sugimoto, K. Astrom, and M. Okutomi.
\newblock Revisiting the pnp problem: A fast, general and optimal solution.
\newblock In \emph{ICCV}, 2013.

\bibitem[Zhou et~al.(2019)Zhou, Barnes, Lu, Yang, and Li]{zhou2019continuity}
Yi Zhou, Connelly Barnes, Jingwan Lu, Jimei Yang, and Hao Li.
\newblock On the continuity of rotation representations in neural networks.
\newblock In \emph{Proceedings of the IEEE/CVF Conference on Computer Vision and Pattern Recognition}, pages 5745--5753, 2019.

\end{thebibliography}
}

\clearpage
\setcounter{page}{1}
\maketitlesupplementary
This supplementary presents the results not shown in the main paper for the lack of space.

\section{GRPS Numerical Stability Full Results}
In addition to rotation errors, we also present the translation and scale error distributions in the following. We can observe consistent performance on both translation and scale errors similar to the rotation errors.

\begin{figure*}[!t]
      \centering
      \begin{subfigure}{0.32\textwidth}
        \includegraphics[width=1.0\linewidth]{figures/figs/R_density.pdf} 
        \caption{Rotation Error (deg)}
        \label{fig:grps_nst}
      \end{subfigure}%
      \begin{subfigure}{0.32\textwidth}
        \includegraphics[width=1.0\linewidth]{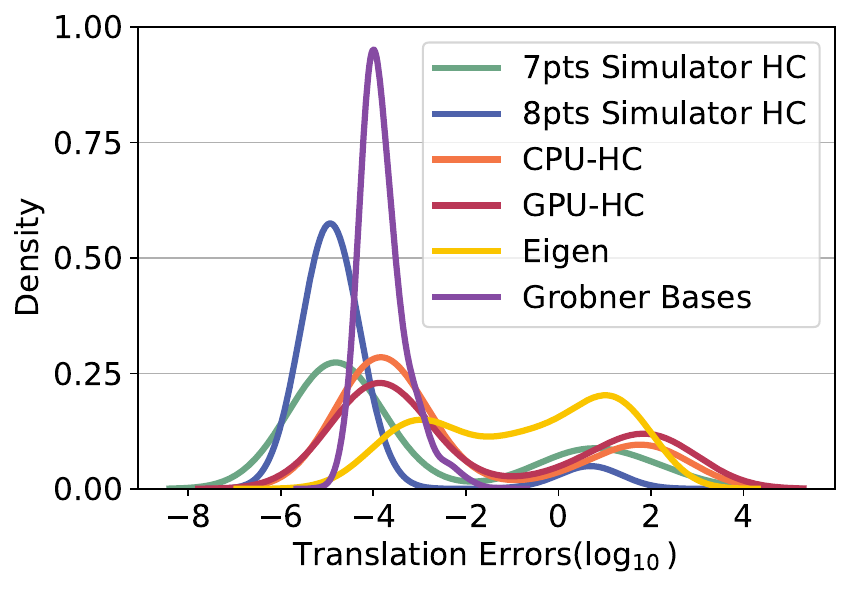} 
        \caption{Translation Error (\%)}
        \label{fig:grps_nst}
      \end{subfigure}%
      \begin{subfigure}{0.32\textwidth}
        \includegraphics[width=1.0\linewidth]{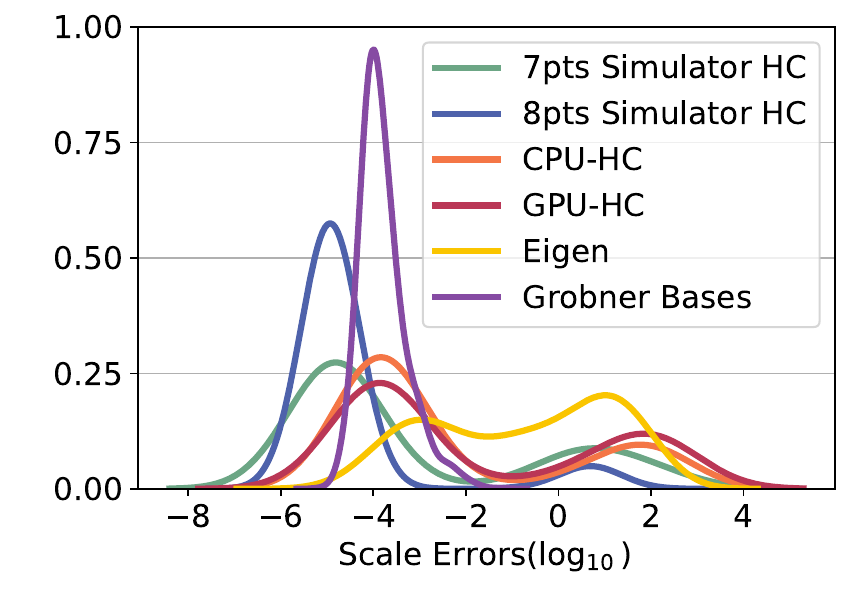} 
        \caption{Scale Error (\%)}
        \label{fig:grps_nss}
      \end{subfigure}
      \begin{minipage}{0.9\textwidth}
      \caption{
      Error distribution over $1000$ trials on noise-free data. The camera number is set to be $3$. Considering the rotation error, except for Gr\"obner Bases, the other minimal solvers all present a similar error distribution with around $70\%$ success rate, and the eigen solver presents only around $60\%$ success rate. The proposed $8$pts simulator HC presents $96.3\%$ success rate where the Gr\"obner Bases has $100\%$ success rate.}
      \end{minipage}
      \vspace{-0.6cm}
      \label{fig:numerical_stability_supp}
\end{figure*}

\section{GRPS Noise Resilience Full Results}
We further present the translation and scale errors for our HC simulator in the noise resilience experiments. Similar performances can be observed on rotation, translation, and scale when varying the noise level. And the rotation errors appear to be less sensitive to image noise than translation and scale errors.

\begin{figure*}[t]
  \centering
  \begin{subfigure}{0.33\textwidth} 
    \includegraphics[width=1.0\linewidth]{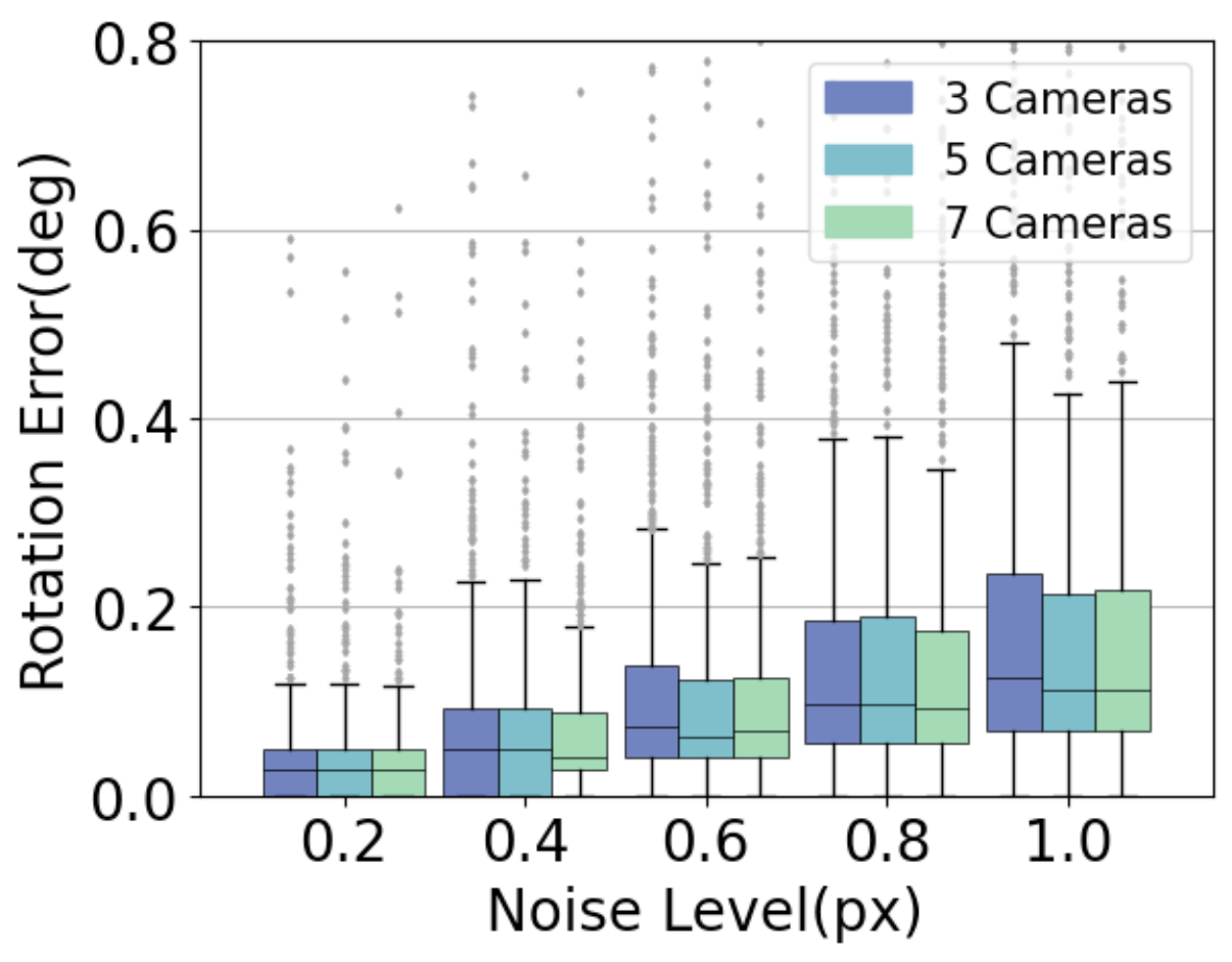} 
    \caption{Rotation Error}
    \label{fig:grps_noisyR}
  \end{subfigure}%
  \begin{subfigure}{0.33\textwidth}
    \includegraphics[width=1.0\linewidth]{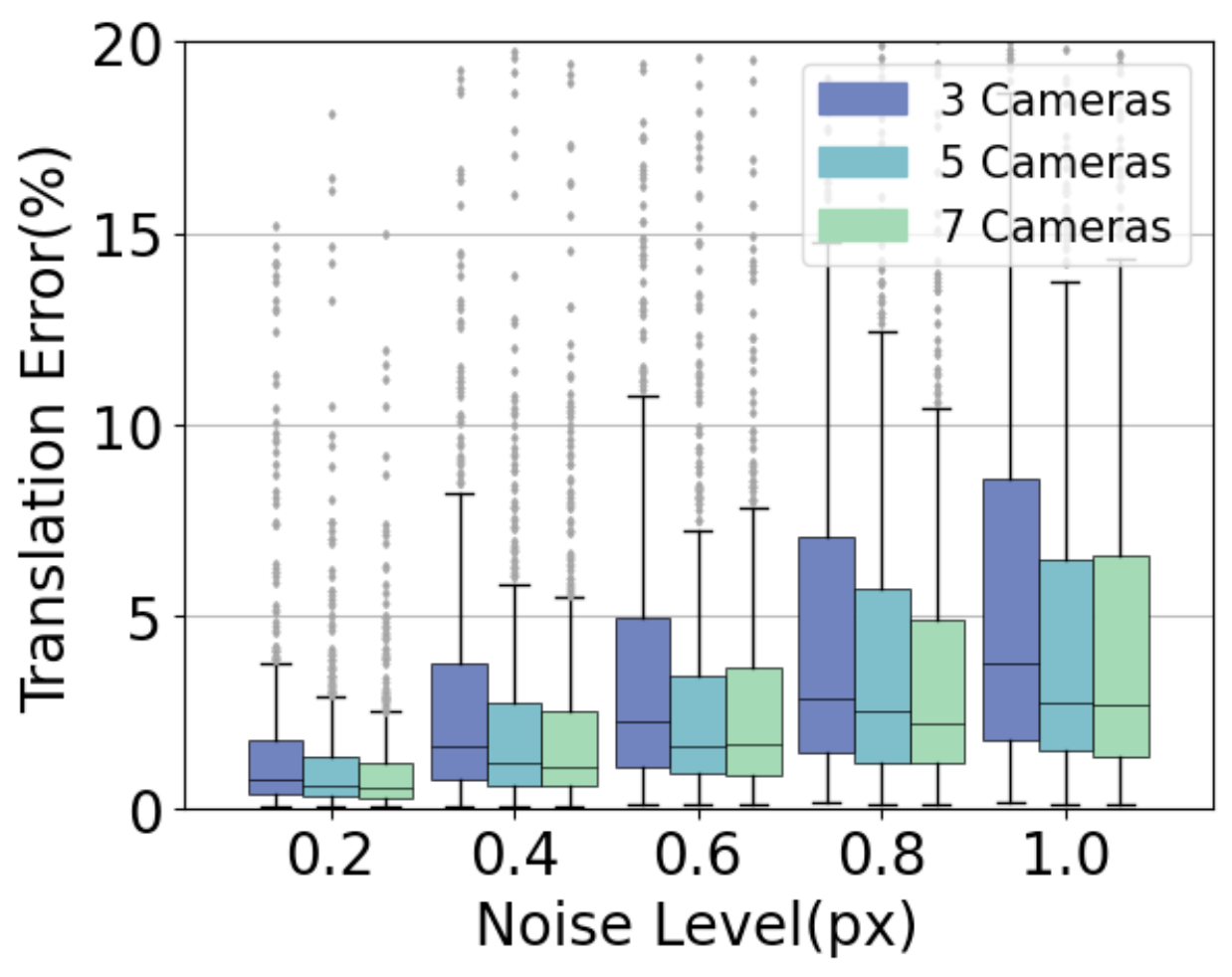} 
    \caption{Translation Error}
    \label{fig:grps_noisyt}
  \end{subfigure}%
  \begin{subfigure}{0.33\textwidth}
    \includegraphics[width=1.0\linewidth]{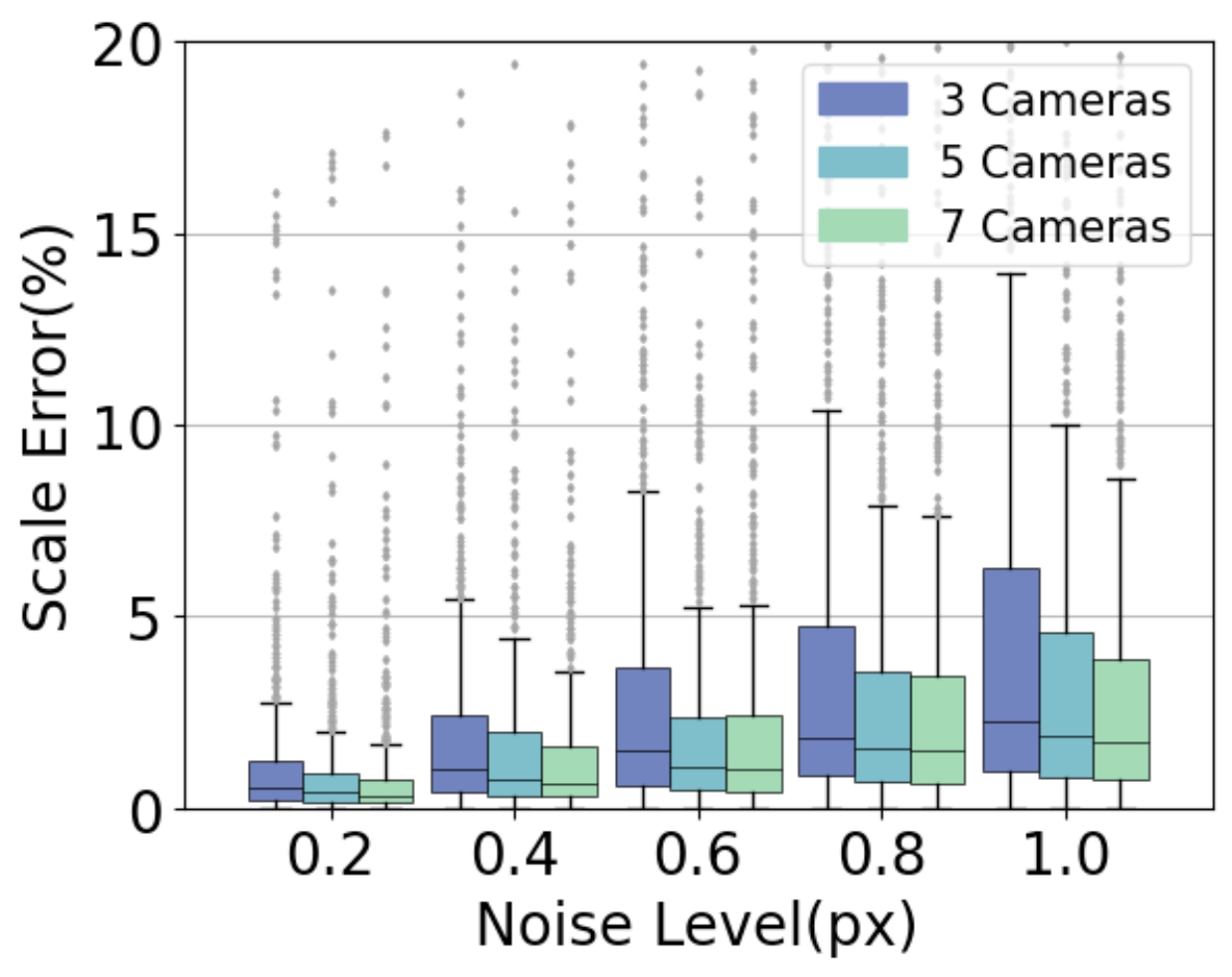} 
    \caption{Scale Error}
    \label{fig:grps_noisys}
  \end{subfigure}
  \begin{minipage}{0.9\textwidth}
  \caption{Error statistics of simulator HC with respect to different noise levels ranging from $0.2$ pixel to $1.0$ pixel.}
      \end{minipage}
  \label{fig:GRPS_noisy_supp}
\end{figure*}

\section{Evaluation within RANSAC}
We further compare the proposed $8$pts simulator HC and Gr\"{o}bner Bases solver inside RANSAC using synthetic data. The following table presents translation error and scale error in percentage.

\begin{table*}[ht]
\centering
\caption{RANSAC experiments for the GRPS problem using Gr\"{o}bner Bases $\mid$ simulator HC. Image points are randomly perturbed by 2-pixel uniform noise. We use vanilla RANSAC with at most $200$ iterations. As can be observed, outlier ratios of up to $40\%$ are easily handled within 200 iterations. And regressor-based simulator HC is consistently faster than Gr\"{o}bner Bases.} 
\begin{tabular}{c|cccccc}
\hline \hline
Ratio & Succ. (\%) & $\mathcal{E}_{\mathbf{R}}(deg)$ &  $\mathcal{E}_{\mathbf{t}}(\%)$ &  $\mathcal{E}_{\mathbf{s}}(\%)$ &  \# Iters & \text{Time(s)}\\ 
\hline
10\% & \;\,99.5 $\mid$ 100 & 0.10 $\mid$ 0.09 & 2.19 $\mid$ 2.03 & 2.13 $\mid$ 1.98 & \;\,93 $\mid$ 125 & 2.33 $\mid$ 0.85\\
20\% & \;\,99.4 $\mid$ 100 & 0.11 $\mid$ 0.10 & 2.31 $\mid$ 2.11 & 2.40 $\mid$ 2.23 & 171 $\mid$ 193 & 4.37 $\mid$ 1.31\\
30\% & \;\,99.4 $\mid$ 100 & 0.13 $\mid$ 0.14 & 2.75 $\mid$ 3.40 & 2.61 $\mid$ 3.17 & 200 $\mid$ 200 & 5.10 $\mid$ 1.35 \\
40\% & 98.9 $\mid$ 91 & 0.21 $\mid$ 0.29 & 5.26 $\mid$ 7.57 & 5.07 $\mid$ 7.81 & 200 $\mid$ 200 & 4.59 $\mid$ 1.35\\
\hline \hline
\end{tabular}
\end{table*}

\section{More Discussion}
We propose a general method to solve nonlinear equations, particularly polynomial systems that challenge traditional approaches such as Gr\"{o}bner Bases (GB) and Levenberg-Marquardt (LM). Given the limitations of classical methods, researchers have actively explored modern alternatives, such as homotopy continuation (HC) and learning-based approaches. However, these methods either induce high computational cost (HC exhaustively tracks all roots) or suffer from poor performance.  
Simulator HC significantly enhances the performance of simple learning-based methods while avoiding the computational expenses of tracking all solution paths, as in standard HC.  

The proposed Simulator HC could be applied to many applications. Examples of important problems for which no alternative exists and where our approach could be applied are given by:
\textit{Rolling Shutter Camera Relative Pose Problem (no minimal solver available)}, \textit{Inverse Kinematics Problem (high degrees of freedom)}. Applying simulator HC to these unsolved problems is a very interesting direction for future research.

\end{document}